\documentclass[10pt,twocolumn,letterpaper]{article}

\usepackage[T1]{fontenc}
\usepackage[utf8]{inputenc}
\usepackage{lmodern}
\usepackage[letterpaper,top=0.8in,bottom=0.8in,left=0.5in,right=0.5in,columnsep=0.25in]{geometry}
\usepackage{graphicx}
\usepackage{amsmath,amssymb}
\usepackage{booktabs}
\usepackage{multirow}
\usepackage[round,authoryear]{natbib}
\usepackage{caption}
\usepackage{placeins}
\usepackage{dblfloatfix}
\usepackage{microtype}
\usepackage[hyphens]{url}
\usepackage[pdfusetitle]{hyperref}

\hypersetup{
  colorlinks=true,
  linkcolor=blue,
  citecolor=blue,
  urlcolor=blue
}
\title{CDSeg: A Renderable Gaussian Carrier for Image-to-3D Label Transfer}

\author{%
  Wentao Sun$^{1}$ \quad
  Yiping Chen$^{2}$ \quad
  Zhengsen Xu$^{3}$ \quad
  Jonathan Li$^{1}$ \quad
  John S. Zelek$^{1}$\\[0.4em]
  \small $^{1}$Department of Systems Design Engineering, University of Waterloo, N2L 3G1 Waterloo, Canada\\
  \small $^{2}$School of Geospatial Engineering and Science, Sun Yat-sen University, 519082 Zhuhai, China\\
  \small $^{3}$Department of Geomatics Engineering, University of Calgary, T2N 1N4 Calgary, Canada\\[0.3em]
  \small\texttt{wentao.sun@uwaterloo.ca, chenyp79@mail.sysu.edu.cn}\\
  \small\texttt{zhengsen.xu@ucalgary.ca, junli@uwaterloo.ca, jzelek@uwaterloo.ca}
}
\date{}

\begin{document}

\maketitle

\begin{abstract}
Modern image models provide strong cues about \emph{what} should be segmented in each view, but their masks do not by themselves determine \emph{where} those labels should persist in 3D. We present Cross-Domain Segmentation via Gaussian Splatting (CDSeg), a label-transfer interface that requires no task-specific 3D segmentation training and uses Gaussian primitives as a renderable label carrier. An external mask source supplies the labels, while renderer-derived visibility determines which 3D primitives receive them. The carrier is instantiated either by completing each input point into one Gaussian, preserving its index, or by reusing the native primitives of an optimized Gaussian scene. CDSeg records pixel--primitive associations during rendering and fuses multi-view masks through voting and a local filter. The resulting labels can be returned to the original points, retained on the native Gaussian scene, or rendered into other views. CDSeg covers promptable, automatic instance, semantic, and LiDAR settings and processes scenes with millions of primitives in seconds. It obtains 92.35\% mIoU on DesktopObjects-360, 95.89\% on NeRDS-360, and 65.77\% on the full ScanNet-v2 validation split using the provided 2D semantic annotations. CDSeg thereby provides one interface for reusing 2D masks across point clouds, Gaussian scenes, and image views without a task-specific 3D segmentation network.
\end{abstract}
\noindent\href{https://w27sun.github.io/CDSeg/}{\textbf{Project page}}
\section{Introduction}
\label{sec:intro}

Dense 3D annotation remains expensive, while modern image models already provide semantic, instance, and promptable masks. Yet these outputs are organized by image view: a mask specifies \emph{what} should be segmented in an image, not \emph{where} that label should persist in 3D. With calibrated views, a shared 3D interface can reuse such masks without training a new point-cloud network for every task.

Direct projection enables this reuse by assigning visible image labels to points and consolidating them across views, but normally terminates at labels on a fixed point set. This does not provide a shared label state when the scene is stored as Gaussian primitives or when the fused labels must be queried from another camera. Recent semantic Gaussian methods attach features or identities to reconstructed scenes~\cite{feature3dgs,gaussian_grouping,property-lang,property-lang2}. FlashSplat, LUDVIG, and PointGS likewise transfer masks or visual features through Gaussian representations~\cite{flashsplat,ludvig,pointgs}, using different solvers, diffusion, or semantic distillation. We instead seek one lightweight label-transfer interface that works with either ordinary points or an existing Gaussian scene, preserves the required 3D identity, and leaves the result renderable.

Gaussian Splatting provides this combination. Its primitives are explicit 3D elements, while its renderer models projected support, depth order, and accumulated transmittance. A primitive can therefore retain a 3D label, and the rendering pass can identify the visible elements associated with each pixel. The Gaussian representation consequently acts as a carrier through which view-wise masks are written into 3D and the fused labels can be read in point, Gaussian, or image space.

\begin{figure*}[t]
    \centering
    \includegraphics[width=\linewidth]{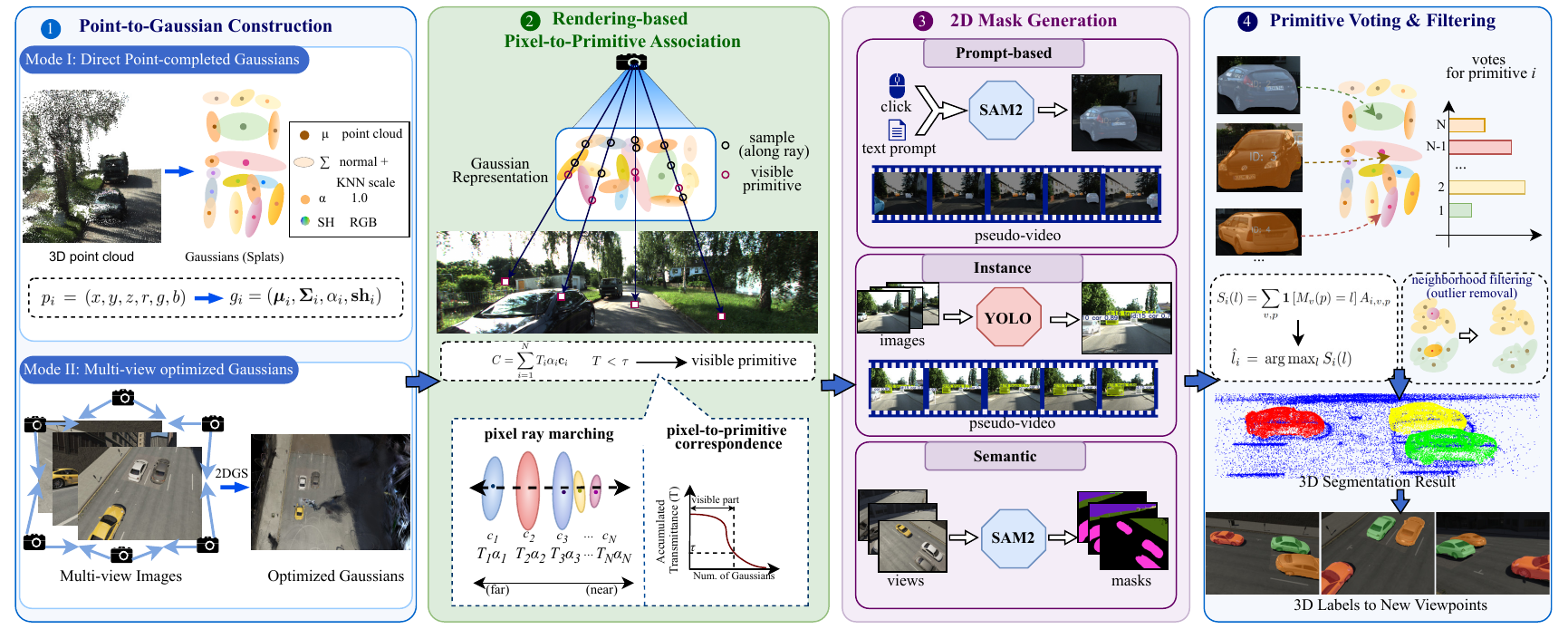}
    \caption{CDSeg connects image masks and 3D labels through a renderable Gaussian carrier. Mode I completes an input point cloud with one Gaussian per point; Mode II reuses an optimized Gaussian scene. The renderer associates pixels with visible primitives, mask labels are fused on those primitives, and the completed labels can be rendered from a target camera.}
    \label{fig:structure}
\end{figure*}

We introduce \textbf{Cross-Domain Segmentation via Gaussian Splatting (CDSeg)}, illustrated in Figure~\ref{fig:structure}. CDSeg separates the two roles: an external mask source determines what is labeled, while Gaussian rendering determines where the label is attached. For point-based tasks, Mode I completes each input point into one Gaussian and preserves its index. For an existing Gaussian scene, Mode II works directly on its optimized primitives. Both modes use the same renderer-derived pixel--primitive association, multi-view voting, and local filtering. The fused labels are then copied back to the supplied points or retained on the native Gaussian carrier, which can also render them from a target camera.

The experiments instantiate this interface in promptable, automatic-instance, semantic, and RGB-to-LiDAR transfer settings. Across tabletop, indoor, outdoor, and driving scenes, they test native-scene labeling, exact point-index compatibility, view coverage, and scaling to millions of primitives.
Our contributions are:
\begin{itemize}
    \item We introduce CDSeg, a label-transfer interface that uses Gaussian primitives to turn view-wise 2D masks into persistent 3D labels without task-specific 3D segmentation training.
    \item We realize the interface in an index-preserving point mode and a native Gaussian-scene mode, with a shared renderer-derived association and fusion procedure that keeps the fused labels available in 3D and image space.
    \item We evaluate both modes across four datasets covering promptable, automatic instance, semantic, and LiDAR transfer, together with view-coverage and million-primitive scaling analyses.
\end{itemize}
\section{Related Work}

\subsection{From Image Masks to 3D Labels}

Point-cloud segmentation is commonly approached with point, voxel, or projection-based models~\cite{pcsurvey}. PointNet++~\cite{point1} builds hierarchical point features; Point Transformer V2/V3~\cite{point4,point2} use attention over local or serialized neighborhoods; sparse voxel methods apply 3D convolutions on a regularized representation~\cite{voxel1,voxel3,voxel6}. These models produce strong task-specific predictors but require 3D training labels.

When calibrated images are available, an alternative is to obtain masks from an image model and transfer them to visible 3D elements. SAM/SAM2~\cite{sam,sam2}, CLIP~\cite{clip}, DINOv2~\cite{dinov2}, and object detectors provide reusable image priors. Direct projection and multi-view voting retain the original point indices and need no 3D network, but the fused result remains a point-based representation. CDSeg uses this point-compatible case as Mode I and extends the same interface to native Gaussian scenes in Mode II.

\subsection{Semantic Gaussian Interfaces}

3D Gaussian Splatting represents a scene with explicit anisotropic primitives and renders them by ordered alpha compositing~\cite{3dgs}; 2DGS improves surface consistency with oriented disks~\cite{2dgs}. Semantic Gaussian methods attach learned features or identities for grouping, open-vocabulary queries, and point-guided segmentation~\cite{feature3dgs,gaussian_grouping,property-lang,property-lang2,pointgauss}.

Several recent methods transfer 2D information to Gaussian primitives. FlashSplat~\cite{flashsplat} solves Gaussian labels from 2D masks with a global linear solver. LUDVIG~\cite{ludvig} aggregates 2D features and refines them by graph diffusion. PointGS~\cite{pointgs} reconstructs a dense Gaussian space from a point cloud, distills SAM masks by contrastive learning, and registers the labeled Gaussians back to the input points. CDSeg instead uses one renderer-derived association and discrete voting procedure for either a one-to-one point-completed carrier or an existing optimized scene. Table~\ref{tab:method_positioning} summarizes these functional differences.

\begin{table*}[t]
\centering
{
\small
\setlength{\tabcolsep}{3pt}
\begin{tabular}{@{}lp{0.18\textwidth}p{0.27\textwidth}cc@{}}
\toprule
Method & Carrier & Transfer & Point index & Gaussian output \\
\midrule
Direct projection & input points & projection and voting & Yes & No \\
FlashSplat~\cite{flashsplat} & optimized Gaussians & global linear solver & No & Yes \\
LUDVIG~\cite{ludvig} & optimized Gaussians & aggregation and graph diffusion & No & Yes \\
PointGS~\cite{pointgs} & reconstructed Gaussians & contrastive distillation and registration & No & Yes \\
CDSeg & points or optimized Gaussians & renderer association and voting & Mode I & Yes \\
\bottomrule
\end{tabular}
}
\caption{Functional positioning of image-to-3D label-transfer interfaces. ``Exact point index'' means that labels return to supplied points without registration or nearest-neighbor remapping. The table compares outputs and transfer mechanisms, not benchmark accuracy.}
\label{tab:method_positioning}
\end{table*}
\section{Method}
\label{sec:method}

Let $\Omega_v$ be the pixel domain of calibrated image $I_v$, and let $M_v:\Omega_v\rightarrow\mathcal{L}\cup\{\varnothing\}$ be its mask map over label set $\mathcal{L}$. Given $\{I_v,M_v\}_{v=1}^{V}$ and a 3D representation, CDSeg constructs a carrier $\mathcal{G}=\{g_i\}_{i=1}^{N}$ and returns a label $\hat{l}_i$ for every observed primitive. Table~\ref{tab:procedure} summarizes the inference path. Carrier construction differs between the two modes; association, fusion, and output are shared.

\begin{table}[t]
\centering
{
\small
\setlength{\tabcolsep}{3pt}
\begin{tabular}{@{}rp{0.78\columnwidth}@{}}
\toprule
Input & Calibrated views, mask maps, and either a point cloud or an optimized Gaussian scene. \\
\midrule
1 & Complete one Gaussian per point (Mode I), or load the native scene primitives (Mode II). \\
2 & Render each view and record its visible pixel--primitive associations. \\
3 & Accumulate mask-label votes on each primitive and select the strongest label. \\
4 & Apply the local label filter; in Mode I, copy labels to the paired input points. \\
\midrule
Output & Labeled points or Gaussian primitives, with optional mapping of the completed 3D labels to image space. \\
\bottomrule
\end{tabular}
}
\caption{Procedure shared by both CDSeg carrier modes.}
\label{tab:procedure}
\end{table}

\subsection{Unified Gaussian Carrier}

A Gaussian primitive is represented as
\begin{equation}
g_i = (\boldsymbol{\mu}_i,\boldsymbol{\Sigma}_i,
       \alpha_i,\mathbf{sh}_i) \in \mathcal{G},
\label{eq:primitive}
\end{equation}
where $\boldsymbol{\mu}_i \in \mathbb{R}^3$ denotes the 3D center, $\boldsymbol{\Sigma}_i \in \mathbb{R}^{3\times3}$ is the covariance matrix encoding the Gaussian's orientation and spatial extent, $\alpha_i \in [0,1]$ represents opacity, and $\mathbf{sh}_i \in \mathbb{R}^{d_{\mathrm{sh}}}$ contains the spherical harmonics coefficients for appearance. CDSeg obtains these primitives in two ways.

\paragraph{Mode I: direct point-completed Gaussians.}
When labels are attached to an input point cloud $\mathcal{P}=\{(\mathbf{x}_i,\mathbf{c}_i)\}_{i=1}^{N}$, every point is completed with the parameters required by Gaussian Splatting. Let $\mathcal{N}_k(i)$ denote its $k$ nearest neighbors, $\mathbf{n}_i$ its estimated normal, $\mathcal{C}(\mathbf{n},s)$ the covariance construction from an orientation and scale, and $\mathcal{H}(\mathbf{c})$ the color-to-SH projection. Mode I is then
\begin{equation}
\begin{aligned}
s_i &= \frac{1}{k}\sum_{j\in\mathcal{N}_k(i)}
       \lVert\mathbf{x}_i-\mathbf{x}_j\rVert_2, \\
g_i^{\mathrm{I}} &=
\bigl(\mathbf{x}_i,\mathcal{C}(\mathbf{n}_i,s_i),
      1,\mathcal{H}(\mathbf{c}_i)\bigr).
\end{aligned}
\label{eq:mode-one}
\end{equation}
No radiance-field optimization, densification, or pruning is performed. The mapping $\mathbf{x}_i\leftrightarrow g_i^{\mathrm{I}}$ is bijective and index preserving, so $\hat l(\mathbf{x}_i)=\hat l_i$ without nearest-neighbor matching. This is important when the annotation space is fixed by the supplied points, as in ScanNet and KITTI-360.

\paragraph{Mode II: multi-view optimized Gaussians.}
If an evaluated benchmark provides a reconstructed Gaussian scene $\mathcal{G}_{\mathrm{opt}}$ and a correspondence to its evaluation geometry, Mode II simply sets $\mathcal{G}=\mathcal{G}_{\mathrm{opt}}$. The scene may be released with the benchmark or reconstructed beforehand using 2DGS or a related method. Reconstruction is not an optimization stage of CDSeg and is excluded from the reported segmentation runtime. This mode preserves the learned geometry, opacity, and view-dependent appearance. We use the optimized scenes released with PointGauss~\cite{pointgauss} for DesktopObjects-360. For NeRDS-360~\cite{nerds360}, we independently optimize a 2DGS~\cite{2dgs} representation for each scene for 30,000 iterations using the original calibrated images.

The required output representation determines the carrier used by CDSeg. Benchmarks scored at supplied point indices use Mode I, because its one-to-one construction keeps the evaluation order intact. Experiments that label a reconstructed scene use Mode II and rely on the available mapping only when scores are computed. The subsequent association and voting steps do not depend on this choice.

\subsection{Rendering-Based Pixel--Primitive Association}

For view $v$ and pixel $p$, let $\mathcal{R}_{v,p}$ be the depth-ordered primitives whose projected support overlaps the pixel. Gaussian Splatting renders their color by front-to-back $\alpha$-blending~\cite{3dgs}:
\begin{equation}
\begin{aligned}
\alpha_{i,v}(p)&=1-\exp[-\sigma_{i,v}(p)\delta_{i,v}(p)],\\
T_{i,v}(p)&=\prod_{j\prec_{v,p}i}\bigl(1-\alpha_{j,v}(p)\bigr),\\
w_{i,v}(p)&=T_{i,v}(p)\alpha_{i,v}(p),\\
C_v(p)&=\sum_{i\in\mathcal{R}_{v,p}}w_{i,v}(p)\mathbf{c}_i(v),
\end{aligned}
    \label{eq:alpha-blending}
\end{equation}
where $j\prec_{v,p}i$ means that $j$ precedes $i$ on the ray, $\sigma_{i,v}(p)$ and $\delta_{i,v}(p)$ determine projected opacity, $T_{i,v}(p)$ is accumulated transmittance, and $w_{i,v}(p)$ is the compositing contribution.

During rendering, let $\nu_{i,v}(p)$ denote the transmittance-based visibility score used by the implementation. The binary pixel--primitive association is
\begin{equation}
A_{i,v,p}=\mathbf{1}[i\in\mathcal{R}_{v,p}]\,
            \mathbf{1}[\nu_{i,v}(p)\geq\tau],
\qquad \tau=0.1.
\label{eq:association}
\end{equation}
Because $A$ is recorded inside the rendering pass, it respects projected support, front-to-back order, and occlusion without a learned correspondence network. A foreground pixel contributes its observed label only to primitives for which $A_{i,v,p}=1$.

For point-completed Gaussians, this operation is close to projecting points and resolving their visibility. For an optimized Gaussian scene, it works directly in the representation used for rendering. The latter is useful when the desired output is a labeled Gaussian scene rather than only labels on a point set.

\subsection{Mask Generation and Multi-View Fusion}

CDSeg is agnostic to the source of $M_v$, but cross-view label consistency depends on the task.
\begin{itemize}
    \item \textbf{Promptable segmentation.} A prompt in the first image initializes SAM2~\cite{sam2}. Calibrated views are ordered by camera-center proximity to form a pseudo-video, allowing the video predictor to propagate the same instance identity across views.
    \item \textbf{Automatic instance segmentation.} YOLOv11~\cite{yolo11_ultralytics} provides per-image detections and masks. Tracking across the pseudo-video gives corresponding objects a shared label before 3D fusion.
    \item \textbf{Semantic segmentation.} Each image supplies a categorical label map, either from a semantic model or from available 2D annotations. No instance tracking is needed because the class indices are shared across images.
\end{itemize}
Keeping mask generation outside the 3D method has two advantages. Different tasks can reuse the same lifting procedure, and competing 3D association methods can be evaluated with identical image masks. The ScanNet setting uses the provided 2D semantic annotations specifically to isolate correspondence and fusion from image recognition.

For primitive $g_i$ and candidate label $l$, the accumulated vote score and observation count are
\begin{equation}
\begin{aligned}
S_i(l)&=\sum_{v=1}^{V}\sum_{p\in\Omega_v}
 A_{i,v,p}\mathbf{1}[M_v(p)=l],\\
o_i&=\sum_{l\in\mathcal{L}}S_i(l).
\end{aligned}
    \label{eq:label-fusion}
\end{equation}
The raw majority-vote label is defined explicitly as
\begin{equation}
l_i^{(0)}=
\begin{cases}
\displaystyle\arg\max_{l\in\mathcal{L}}S_i(l), & o_i>0,\\
\varnothing, & o_i=0.
\end{cases}
\label{eq:raw-label}
\end{equation}
Repeated cross-view agreement therefore increases support, while an isolated mask error affects only its contributing views. A primitive with $o_i=0$ remains unlabeled.

After voting, a fixed $k$-nearest-neighbor majority filter suppresses isolated disagreements. With $\bar{\mathcal{N}}_k(i)=\mathcal{N}_k(i)\cup\{i\}$, its output is
\begin{equation}
\hat l_i=
\begin{cases}
\displaystyle\arg\max_{l\in\mathcal{L}}
\sum_{j\in\bar{\mathcal{N}}_k(i)}\mathbf{1}[l_j^{(0)}=l],
& o_i>0,\\
\varnothing, & o_i=0.
\end{cases}
\label{eq:local-filter}
\end{equation}
The support condition prevents the filter from filling a wholly unobserved region. In Mode I, the filtered labels are copied to their index-paired input points.


\paragraph{Optional mapping of 3D labels to images.}
Once CDSeg has obtained the 3D labels, image-space labels can be recovered independently in several ways. A direct option projects the labeled 3D points or primitives through a calibrated camera, with the chosen projection rule resolving visibility. Alternatively, the 3DGS renderer can monitor accumulated transmittance along each pixel ray. Let $T^{+}_{i,v}(p)$ be the transmittance remaining after the ray passes primitive $i$. The surface label is taken from the first depth-ordered primitive that reduces the remaining transmittance below the renderer's stopping threshold $\tau_{\mathrm{stop}}$:
\begin{equation}
\begin{aligned}
T^{+}_{i,v}(p)
  &=T_{i,v}(p)\bigl(1-\alpha_{i,v}(p)\bigr),\\
i_v^\star(p)
  &=\operatorname{first}_{i\in\mathcal{R}_{v,p}}
    \left\{T^{+}_{i,v}(p)\leq\tau_{\mathrm{stop}}\right\},\\
L_v^{\mathrm{surf}}(p)
  &=\hat l_{i_v^\star(p)}.
\end{aligned}
\label{eq:surface-label}
\end{equation}
Here, $\operatorname{first}$ follows the front-to-back order in $\mathcal{R}_{v,p}$; if no primitive crosses the threshold, the pixel remains unlabeled. A third option encodes the 3D labels as display colors. Let $\mathbf{q}:\mathcal{L}\cup\{\varnothing\}\rightarrow\mathbb{R}^3$ with $\mathbf{q}(\varnothing)=\mathbf{0}$. Replacing the original primitive colors with $\mathbf{q}(\hat l_i)$ gives
\begin{equation}
\widetilde C_v(p)=\sum_{i\in\mathcal{R}_{v,p}}
w_{i,v}(p)\mathbf{q}(\hat l_i).
\label{eq:label-rendering}
\end{equation}
These mappings query labels already fused in 3D. They are used only to obtain an image-space output from a calibrated camera and do not change any primitive label.

\paragraph{Relation to direct projection.}
Both direct projection and CDSeg lift multi-view masks without 3D training. Direct projection associates pixels with projected points, optionally resolving visibility with a z-buffer, and terminates with point labels. CDSeg associates each pixel with the Gaussian primitives along its camera ray by recording the transmittance changes induced as the ray traverses them. With one Gaussian completed from each input point and no scene optimization, Mode I reduces to the point-projection case while preserving point indices within the same rendering interface. Mode II applies the association directly to optimized Gaussian scenes and retains the labeled scene for subsequent rendering. Together, the two modes provide a common renderable carrier across points, images, and Gaussian scenes.
\begin{figure*}[t]
    \centering
    \includegraphics[width=\linewidth]{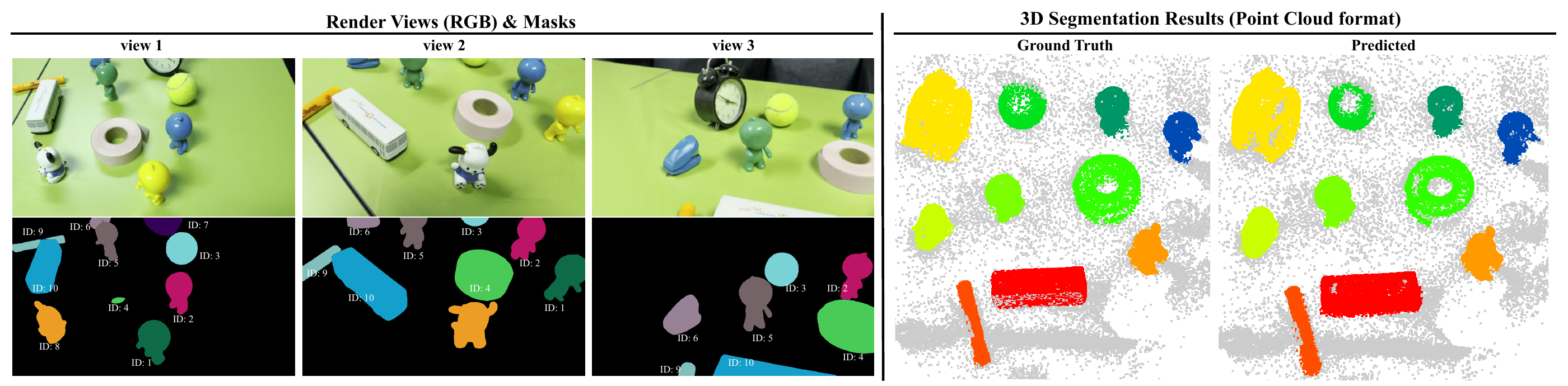}
    \caption{Promptable segmentation on the \emph{Desk 1} scene of DesktopObjects-360. Representative views and propagated SAM2 masks are followed by the CDSeg prediction and ground-truth 3D annotation. Colors identify object instances.}
    \label{fig:prompt}
\end{figure*}
\section{Experiments}
\label{sec:experiments}

We test CDSeg as a reusable image-to-3D interface. Mode II evaluates prompted and automatic-instance labels on native Gaussian scenes; Mode I evaluates exact-index outputs on meshes and LiDAR. Coverage and scaling probe visibility and computational limits.

\subsection{Experimental Protocol}

\paragraph{Datasets and carrier modes.}
DesktopObjects-360, introduced with PointGauss~\cite{pointgauss}, contains six tabletop scenes, 3,364 views, and 56 annotated objects. NeRDS-360~\cite{nerds360} provides the 360-degree outdoor image sequences and calibrated viewpoints. For these scenes, we optimize the Gaussian carriers ourselves using 2DGS~\cite{2dgs} for 30,000 iterations and manually annotate the vehicle instances on the evaluation geometry. These annotations are used only as evaluation ground truth and are not available to CDSeg during segmentation. Both benchmarks use optimized carriers (Mode II). ScanNet~\cite{scannet} provides indoor RGB-D scans, poses, and semantic annotations, while KITTI-360~\cite{kitti360} provides driving images, LiDAR, poses, and point labels. Their evaluation targets are supplied points, so both use point-completed carriers (Mode I).
\begin{table*}[t]
\centering
{
\small
\begin{tabular}{@{}llll@{}}
\toprule
Dataset & Task and 2D input & Carrier & Evaluation target \\
\midrule
DesktopObjects-360 & prompted instances from SAM2 & Mode II: optimized Gaussians & annotated scene objects \\
NeRDS-360 & automatic instances from YOLOv11 & Mode II: our 2DGS (30k iterations) & manually annotated vehicles \\
ScanNet-v2 val & provided coarse 2D semantic annotations & Mode I: point-completed & official benchmark vertices \\
KITTI-360 & vehicle instances from YOLOv11 & Mode I: point-completed & supplied LiDAR points \\
\bottomrule
\end{tabular}
}
\caption{Evaluation protocol and role of each dataset. Mode II tests direct labeling of optimized Gaussian scenes; Mode I tests point-indexed output on meshes and LiDAR. Reconstruction and 2D-mask inference are outside CDSeg.}
\label{tab:protocol}
\end{table*}

\begin{figure*}[t]
    \centering
    \includegraphics[width=\linewidth]{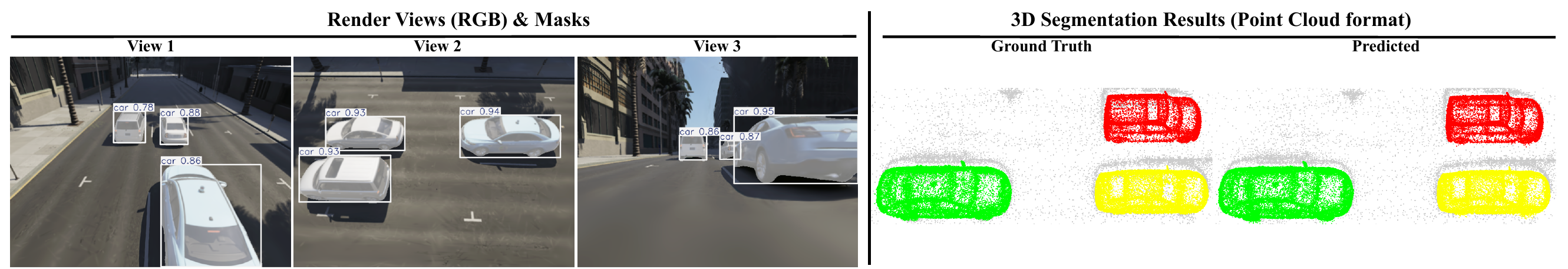}
    \caption{Automatic instance segmentation on the \emph{SF\_VanNessAveAndTurkSt5} scene of NeRDS-360. Representative views contain YOLOv11 masks; the last panels compare the completed CDSeg result with the 3D ground truth.}
    \label{fig:inst_nerds}
\end{figure*}
\paragraph{Tasks and masks.}
DesktopObjects-360 uses a first-view prompt with SAM2 to test native-scene labeling; NeRDS-360 uses YOLOv11 to test automatic multi-instance labeling of optimized Gaussians. KITTI-360 tests transfer to sparse outdoor LiDAR, while ScanNet uses provided 2D annotations to isolate semantic transfer to official benchmark vertices. Table~\ref{tab:protocol} summarizes the settings.

\paragraph{Baselines and metrics.}
PointNeXt~\cite{pointnext}, Point Transformer V2/V3~\cite{point4,point2}, and PointMetaBase~\cite{pointmeta} provide leave-one-scene-out 3D accuracy context for DesktopObjects-360 and NeRDS-360. Projection-and-vote controls, with and without z-buffer visibility, receive the same masks as CDSeg and isolate label transfer from mask generation. ScanNet-v2 references span full, weak, and self-supervised 3D training. We report mean intersection-over-union (mIoU), overall point accuracy (mAcc), mean per-class accuracy (mAcc-cls), and ScanNet global valid-point accuracy (oAcc) where applicable.

\paragraph{Implementation.}
All experiments use an RTX 4090, an i5-11400F, PyTorch 2.5.1, and CUDA 12.4. Reported CDSeg time starts with an available Gaussian carrier and masks, and includes correspondence extraction, voting, and filtering. It excludes off-the-shelf 2D inference and, for Mode II, prior Gaussian reconstruction, because neither is trained or optimized by CDSeg. Supplementary material lists evaluated scenes, mask processing, and extended per-scene results.

\subsection{Coverage and Efficiency}

\begin{table}[t]
\centering
{
\small
\setlength{\tabcolsep}{2pt}
\begin{tabular}{@{}lrrrrr@{}}
\toprule
\multicolumn{6}{c}{\textbf{View coverage on DesktopObjects-360 Desk6}} \\
Input views & 206 & 103 & 41 & 10 & 1 \\
mIoU (\%) & 93.36 & 93.41 & 93.39 & 93.21 & 74.50 \\
\midrule
\multicolumn{6}{c}{\textbf{Scaling with point count}} \\
Points (million) & 1 & 3 & 5 & 7 & -- \\
Runtime (s) & 2.23 & 2.55 & 2.90 & 3.74 & -- \\
GPU memory (GB) & 5.94 & 7.30 & 8.80 & 10.40 & -- \\
\midrule
\multicolumn{6}{c}{\textbf{Scaling with view count}} \\
Input views & 10 & 100 & 200 & 300 & 500 \\
Runtime (s) & 0.21 & 0.89 & 1.50 & 2.23 & 3.30 \\
GPU memory (GB) & 1.16 & 2.72 & 4.34 & 5.94 & 9.04 \\
\bottomrule
\end{tabular}
}
\caption{Coverage and scaling. Runtime excludes external 2D-mask inference and prior Gaussian reconstruction. Point scaling uses 300 views; view scaling uses one million points. Memory is peak GPU memory.}
\label{tab:scaling}
\end{table}

Table~\ref{tab:scaling} separately measures the effect of view coverage and the runtime and memory costs of increasing the numbers of input views and carrier primitives. Desk6 remains above 93.2\% mIoU with ten selected views, but falls to 74.5\% with one view; in this ablation, surface coverage appears more important than dense view count. Runtime and memory grow approximately linearly. With 300 masks, increasing the carrier from one to seven million primitives raises runtime from 2.23 to 3.74 seconds; with one million primitives, 500 views take 3.30 seconds. The association is accumulated during rendering, avoiding storage proportional to every primitive--pixel pair.

\subsection{Promptable Segmentation}

A prompt is supplied only in the first view, and SAM2 propagates the selected object through the pseudo-video. This setting tests whether the resulting multi-view masks can become persistent instance labels on a native Gaussian scene.

\begin{table}[t]
\centering
{
\small
\begin{tabular}{@{}lccc@{}}
\toprule
Method & mIoU & mAcc & mAcc-cls \\
\midrule
PointNeXt~\cite{pointnext} & 87.21 & 98.07 & 87.79 \\
PTv2~\cite{point4} & 90.13 & 95.77 & 91.09 \\
PointMetaBase~\cite{pointmeta} & 90.31 & 95.57 & 89.55 \\
PTv3~\cite{point2} & 91.97 & 95.75 & 93.22 \\
\midrule
Projection + vote & \textbf{92.89} & 96.43 & 95.55 \\
Projection + z-buffer + vote & 91.77 & 95.53 & 94.16 \\
CDSeg & 92.35 & \textbf{96.54} & \textbf{95.62} \\
\bottomrule
\end{tabular}
}
\caption{Promptable segmentation on DesktopObjects-360 (\%). Point networks use 3D supervision; projection variants and CDSeg receive identical SAM2 masks.}
\label{tab:desktop}
\end{table}

\begin{table}[t]
\centering
{
\small
\begin{tabular}{@{}lccc@{}}
\toprule
Method & mIoU & mAcc & mAcc-cls \\
\midrule
PointNeXt~\cite{pointnext} & 89.13 & 98.72 & 90.04 \\
PTv2~\cite{point4} & \textbf{96.29} & \textbf{99.56} & 97.23 \\
PointMetaBase~\cite{pointmeta} & 94.32 & 99.30 & 95.30 \\
PTv3~\cite{point2} & 94.64 & 99.35 & 95.80 \\
\midrule
Projection + vote & 91.82 & 98.92 & 99.73 \\
Projection + z-buffer + vote & 94.67 & 99.34 & 99.57 \\
CDSeg & 95.89 & 99.50 & \textbf{99.86} \\
\bottomrule
\end{tabular}
}
\caption{Automatic instance segmentation on NeRDS-360 (\%). Point networks use 3D supervision; lifting methods receive identical YOLOv11 masks.}
\label{tab:nerds}
\end{table}

On DesktopObjects-360, CDSeg reaches 92.35\% mIoU and gives the highest mAcc and mAcc-cls in Table~\ref{tab:desktop}. The labels remain on the optimized Gaussian carrier rather than being reduced to a point-only output.

All methods receive the same propagated masks. Figure~\ref{fig:prompt} shows that the prompted identity is transferred to the corresponding 3D object, including closely arranged objects with challenging contact boundaries.

\subsection{Automatic Instance Segmentation}

Unlike the prompted experiment, this setting begins with automatically detected vehicle masks. YOLOv11 provides masks for multiple vehicles, and tracking assigns their instance IDs across the NeRDS-360 views. We fuse these masks on independently optimized 2DGS scenes. CDSeg obtains 95.89\% mIoU, 99.50\% mAcc, and 99.86\% mAcc-cls without 3D training supervision (Table~\ref{tab:nerds}). The manually annotated vehicle instances are withheld from CDSeg and used only for evaluation. Figure~\ref{fig:inst_nerds} shows that identities remain consistent across views and align with the annotated objects.

The projection baselines and CDSeg receive the same tracked masks, so their comparison isolates the 3D association and fusion stages. All predictions are scored on the same evaluation points. The projection baselines label these points directly, whereas CDSeg maps the labels stored on the optimized NeRDS-360 primitives to them for evaluation. The labels remain on the primitives and can also be rendered from calibrated views. Mode I is evaluated separately in the ScanNet and KITTI-360 experiments. 


\begin{figure*}[t]
    \centering
    \includegraphics[width=\linewidth]{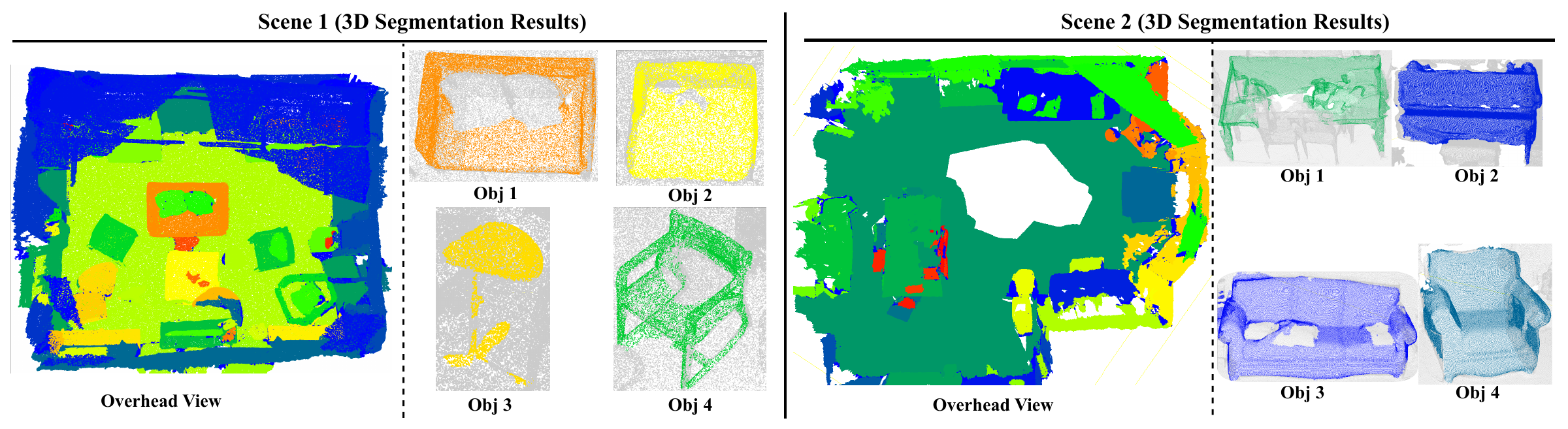}
    \caption{Semantic-mask lifting on ScanNet. The figure shows four completed indoor scenes and enlarged regions containing representative furniture categories. Colors denote the 20 semantic classes.}
    \label{fig:sem_scannet}
\end{figure*}

\begin{figure*}[t]
    \centering
    \includegraphics[width=\linewidth]{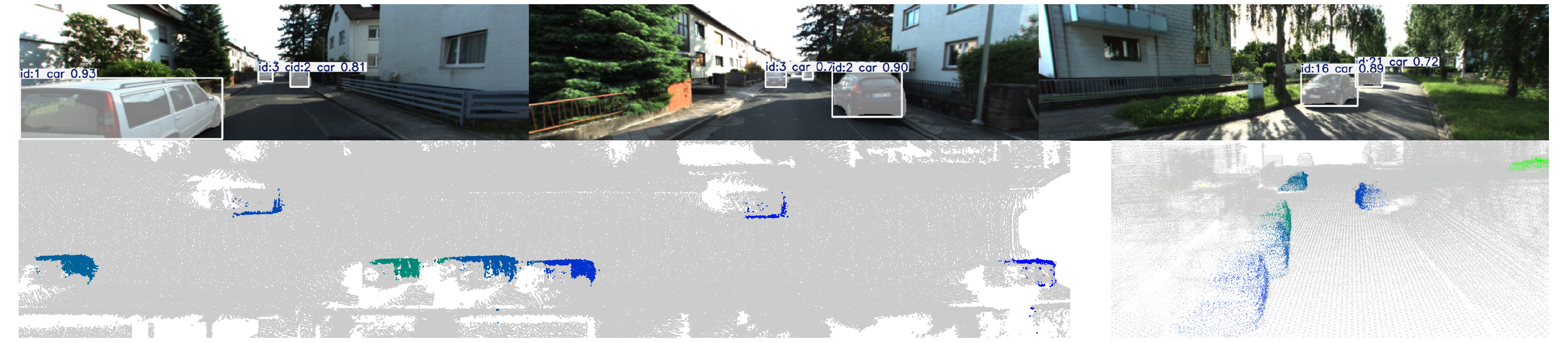}
    \caption{KITTI-360 vehicle segmentation. The upper row shows image-space YOLOv11 masks and the lower row shows their corresponding labels on the accumulated LiDAR points.}
    \label{fig:inst_kitti}
\end{figure*}
\subsection{Semantic-Mask Lifting and LiDAR Transfer}

\paragraph{ScanNet semantic-mask lifting.}
ScanNet isolates 2D-to-3D semantic association from image recognition: the provided 2D annotations serve as controlled masks, while the corresponding 3D properties are withheld for evaluation. We use the full validation split. Because Mode I retains every benchmark vertex index, predictions are scored directly without geometric remapping.

\begin{table}[t]
\centering
{
\footnotesize
\setlength{\tabcolsep}{1.2pt}
\newcommand{\scmethod}[2]{\parbox[t]{0.46\columnwidth}{\raggedright #1~\cite{#2}\par}}
\begin{tabular}{@{}lcccc@{}}
\toprule
Method & mIoU & \shortstack{Large-\\Scale} & \shortstack{3D\\Supervision} & \shortstack{Zero-\\Shot} \\
\midrule
\scmethod{RandLA-Net}{randla} & 64.5 & Yes & Full & No \\
\scmethod{One Thing One Click}{onething1click} & 69.3 & No & Weak & No \\
\scmethod{PointContrast}{pointcontrast} & 68.3 & No & Self-sup. & No \\
\scmethod{ContrastBoundary}{contrastiveboundary} & 70.5 & No & Self-sup. & No \\
\scmethod{3DSS-VLG}{3DSS-VLG} & 48.9 & No & Weak & No \\
\midrule
\textbf{CDSeg (Ours)} & 65.77 & \textbf{Yes} & \textbf{None} & \textbf{Yes} \\
\bottomrule
\end{tabular}
}
\caption{ScanNet-v2 validation comparison (\%). Self-sup. denotes self-supervised pretraining; supervision refers to 3D training.}
\label{tab:scannet_baselines}
\end{table}

Using the provided 20-class 2D annotations, CDSeg obtains 65.77\% global-confusion mIoU and 73.37\% oAcc on the full validation split without 3D training. In Table~\ref{tab:scannet_baselines}, it exceeds large-scale RandLA-Net (64.5\% mIoU) but trails the strongest weakly or self-supervised methods. CDSeg alone is marked both large-scale and zero-shot with no 3D supervision; the provided annotations are image inputs. The supplement reports class-level results.

Figure~\ref{fig:sem_scannet} shows that large planar structures remain coherent across the completed point cloud and that common furniture classes retain recognizable boundaries. The enlarged regions further demonstrate detailed label transfer around object contacts and compact furniture.

\paragraph{KITTI-360 LiDAR transfer.}
This setting tests point-indexed transfer to sparse outdoor geometry: YOLOv11 masks from calibrated RGB images are fused on the point-completed LiDAR carrier without reconstructing a separate Gaussian scene. We evaluate stationary vehicles in ten sequence-0 windows, and Mode I preserves all accumulated LiDAR evaluation points.

CDSeg obtains 57.44\% mIoU and 72.87\% F1, with 83.77\% precision and 64.48\% recall. The high precision demonstrates reliable label transfer on camera-visible vehicle geometry across the evaluated driving windows. Figure~\ref{fig:inst_kitti} shows stable vehicle labels across multiple images and their consistent transfer to the accumulated LiDAR geometry.

\paragraph{Mapping labels back to images.}
After fusion, the completed 3D labels can be mapped to image space by direct projection, by assigning the label of the first Gaussian that exhausts accumulated transmittance along each pixel ray, or by replacing primitive colors with label colors and invoking the 3DGS compositing pipeline. These output choices do not alter the 3D fusion result. Further details on mapping the 3D labels back to images are provided in the supplementary material.

\subsection{Discussion}

Results position CDSeg as a reusable label-transfer interface: Mode I preserves exact point indices, whereas Mode II keeps fused labels on native renderable primitives. Both avoid task-specific 3D training but inherit the coverage and quality of their input masks and carrier. Their value is interoperability rather than universal numerical superiority.

\subsection{Limitations}

CDSeg transfers the evidence present in its input masks, so a surface absent from every view remains unlabeled. Mode II also assumes an optimized scene and a correspondence to the evaluation geometry. Consequently, if the Gaussian scene itself contains defects, such as missing regions or distortions, CDSeg will also be affected.

\section{Conclusion}

CDSeg turns view-wise masks into reusable 3D labels through a renderable Gaussian carrier rather than a task-specific 3D model. Mode I preserves point indices; Mode II retains labels on native optimized Gaussians. Both share renderer-derived association and multi-view fusion. Across four datasets, the interface supports prompted, automatic-instance, semantic, and LiDAR transfer without task-specific 3D training and scales to millions of primitives. Future work will improve Mode I to preserve finer texture details during point-cloud-to-Gaussian conversion.

\FloatBarrier
\bibliographystyle{plainnat}
\bibliography{main}

\begin{thebibliography}{32}
\providecommand{\natexlab}[1]{#1}
\providecommand{\url}[1]{\texttt{#1}}
\expandafter\ifx\csname urlstyle\endcsname\relax
  \providecommand{\doi}[1]{doi: #1}\else
  \providecommand{\doi}{doi: \begingroup \urlstyle{rm}\Url}\fi

\bibitem[Chen et~al.(2023)Chen, Xu, Chen, Zhou, Xiao, Sun, Xie, and
  Kang]{voxel6}
Zisheng Chen, Hongbin Xu, Weitao Chen, Zhipeng Zhou, Haihong Xiao, Baigui Sun,
  Xuansong Xie, and Wenxiong Kang.
\newblock Pointdc: Unsupervised semantic segmentation of {3D} point clouds via
  cross-modal distillation and super-voxel clustering.
\newblock In \emph{Proceedings of the IEEE/CVF International Conference on
  Computer Vision (ICCV)}, pages 14290--14299, October 2023.

\bibitem[Dai et~al.(2017)Dai, Chang, Savva, Halber, Funkhouser, and
  Nie{\ss}ner]{scannet}
Angela Dai, Angel~X. Chang, Manolis Savva, Maciej Halber, Thomas Funkhouser,
  and Matthias Nie{\ss}ner.
\newblock Scannet: Richly-annotated 3d reconstructions of indoor scenes.
\newblock In \emph{Proceedings of the IEEE Conference on Computer Vision and
  Pattern Recognition (CVPR)}, pages 2432--2443, 2017.
\newblock \doi{10.1109/CVPR.2017.261}.

\bibitem[Dai et~al.(2018)Dai, Ritchie, Bokeloh, Reed, Sturm, and
  Nießner]{voxel1}
Angela Dai, Daniel Ritchie, Martin Bokeloh, Scott Reed, Jürgen Sturm, and
  Matthias Nießner.
\newblock Scancomplete: Large-scale scene completion and semantic segmentation
  for 3d scans.
\newblock In \emph{2018 IEEE/CVF Conference on Computer Vision and Pattern
  Recognition}, pages 4578--4587, 2018.
\newblock \doi{10.1109/CVPR.2018.00481}.

\bibitem[Guo et~al.(2021)Guo, Wang, Hu, Liu, Liu, and Bennamoun]{pcsurvey}
Yulan Guo, Hanyun Wang, Qingyong Hu, Hao Liu, Li~Liu, and Mohammed Bennamoun.
\newblock Deep learning for 3d point clouds: A survey.
\newblock \emph{IEEE Transactions on Pattern Analysis and Machine
  Intelligence}, 43\penalty0 (12):\penalty0 4338--4364, 2021.
\newblock \doi{10.1109/TPAMI.2020.3005434}.

\bibitem[Hu et~al.(2020)Hu, Yang, Xie, Rosa, Guo, Wang, Trigoni, and
  Markham]{randla}
Qingyong Hu, Bo~Yang, Linhai Xie, Stefano Rosa, Yulan Guo, Zhihua Wang, Niki
  Trigoni, and Andrew Markham.
\newblock Randla-net: Efficient semantic segmentation of large-scale point
  clouds.
\newblock In \emph{Proceedings of the IEEE/CVF conference on computer vision
  and pattern recognition}, pages 11108--11117, 2020.

\bibitem[Huang et~al.(2024)Huang, Yu, Chen, Geiger, and Gao]{2dgs}
Binbin Huang, Zehao Yu, Anpei Chen, Andreas Geiger, and Shenghua Gao.
\newblock {2D} gaussian splatting for geometrically accurate radiance fields.
\newblock In \emph{ACM SIGGRAPH 2024 Conference Papers}. Association for
  Computing Machinery, 2024.
\newblock \doi{10.1145/3641519.3657428}.

\bibitem[Irshad et~al.(2023)Irshad, Zakharov, Liu, Guizilini, Kollar, Gaidon,
  Kira, and Ambrus]{nerds360}
Muhammad~Zubair Irshad, Sergey Zakharov, Katherine Liu, Vitor Guizilini, Thomas
  Kollar, Adrien Gaidon, Zsolt Kira, and Rares Ambrus.
\newblock Neo 360: Neural fields for sparse view synthesis of outdoor scenes.
\newblock In \emph{2023 IEEE/CVF International Conference on Computer Vision
  (ICCV)}, pages 9187--9198, 2023.
\newblock \doi{10.1109/ICCV51070.2023.00843}.

\bibitem[Jocher and Qiu(2024)]{yolo11_ultralytics}
Glenn Jocher and Jing Qiu.
\newblock Ultralytics yolo11, 2024.
\newblock URL \url{https://github.com/ultralytics/ultralytics}.

\bibitem[Kerbl et~al.(2023)Kerbl, Kopanas, Leimk{\"u}hler, and Drettakis]{3dgs}
Bernhard Kerbl, Georgios Kopanas, Thomas Leimk{\"u}hler, and George Drettakis.
\newblock 3d gaussian splatting for real-time radiance field rendering.
\newblock \emph{ACM Transactions on Graphics}, 42\penalty0 (4), July 2023.
\newblock \doi{10.1145/3592433}.
\newblock URL \url{https://repo-sam.inria.fr/fungraph/3d-gaussian-splatting/}.

\bibitem[Kirillov et~al.(2023)Kirillov, Mintun, Ravi, Mao, Rolland, Gustafson,
  Xiao, Whitehead, Berg, Lo, Dollar, and Girshick]{sam}
Alexander Kirillov, Eric Mintun, Nikhila Ravi, Hanzi Mao, Chloe Rolland, Laura
  Gustafson, Tete Xiao, Spencer Whitehead, Alexander~C. Berg, Wan-Yen Lo, Piotr
  Dollar, and Ross Girshick.
\newblock Segment anything.
\newblock In \emph{Proceedings of the IEEE/CVF International Conference on
  Computer Vision (ICCV)}, pages 4015--4026, October 2023.

\bibitem[Liao et~al.(2023)Liao, Xie, and Geiger]{kitti360}
Yiyi Liao, Jun Xie, and Andreas Geiger.
\newblock Kitti-360: A novel dataset and benchmarks for urban scene
  understanding in 2d and 3d.
\newblock \emph{IEEE Transactions on Pattern Analysis and Machine
  Intelligence}, 45\penalty0 (3):\penalty0 3292--3310, 2023.
\newblock \doi{10.1109/TPAMI.2022.3179507}.

\bibitem[Lin et~al.(2023)Lin, Zheng, Li, Chao, Wang, Wang, Tian, and
  Ji]{pointmeta}
Haojia Lin, Xiawu Zheng, Lijiang Li, Fei Chao, Shanshan Wang, Yan Wang,
  Yonghong Tian, and Rongrong Ji.
\newblock Meta architecture for point cloud analysis.
\newblock In \emph{2023 IEEE/CVF Conference on Computer Vision and Pattern
  Recognition (CVPR)}, pages 17682--17691, 2023.
\newblock \doi{10.1109/CVPR52729.2023.01696}.

\bibitem[Liu et~al.(2021)Liu, Qi, and Fu]{onething1click}
Zhengzhe Liu, Xiaojuan Qi, and Chi-Wing Fu.
\newblock One thing one click: A self-training approach for weakly supervised
  3d semantic segmentation.
\newblock In \emph{Proceedings of the IEEE/CVF conference on computer vision
  and pattern recognition}, pages 1726--1736, 2021.

\bibitem[Marrie et~al.(2025)Marrie, Menegaux, Arbel, Larlus, and
  Mairal]{ludvig}
Juliette Marrie, Romain Menegaux, Michael Arbel, Diane Larlus, and Julien
  Mairal.
\newblock {LUDVIG}: Learning-free uplifting of {2D} visual features to gaussian
  splatting scenes.
\newblock In \emph{Proceedings of the IEEE/CVF International Conference on
  Computer Vision}, pages 7440--7450, 2025.

\bibitem[Oquab et~al.(2024)Oquab, Darcet, Moutakanni, Vo, Szafraniec, Khalidov,
  Fernandez, Haziza, Massa, El-Nouby, Assran, Ballas, Galuba, Howes, Huang, Li,
  Misra, Rabbat, Sharma, Synnaeve, Xu, Jégou, Mairal, Labatut, Joulin, and
  Bojanowski]{dinov2}
Maxime Oquab, Timothée Darcet, Théo Moutakanni, Huy Vo, Marc Szafraniec,
  Vasil Khalidov, Pierre Fernandez, Daniel Haziza, Francisco Massa, Alaaeldin
  El-Nouby, Mido Assran, Nicolas Ballas, Wojciech Galuba, Russell Howes, Po-Yao
  Huang, Shang-Wen Li, Ishan Misra, Michael Rabbat, Vasu Sharma, Gabriel
  Synnaeve, Hu~Xu, Hervé Jégou, Julien Mairal, Patrick Labatut, Armand
  Joulin, and Piotr Bojanowski.
\newblock {DINOv2}: Learning robust visual features without supervision.
\newblock \emph{Transactions on Machine Learning Research}, 2024.
\newblock ISSN 2835-8856.

\bibitem[Park et~al.(2023)Park, Kim, Kim, and Jo]{voxel3}
Jaehyun Park, Chansoo Kim, Soyeong Kim, and Kichun Jo.
\newblock Pcscnet: Fast 3d semantic segmentation of lidar point cloud for
  autonomous car using point convolution and sparse convolution network.
\newblock \emph{Expert Systems with Applications}, 212:\penalty0 118815, 2023.
\newblock ISSN 0957-4174.
\newblock \doi{10.1016/j.eswa.2022.118815}.
\newblock URL
  \url{https://www.sciencedirect.com/science/article/pii/S0957417422018334}.

\bibitem[Qi et~al.(2017)Qi, Yi, Su, and Guibas]{point1}
Charles~R. Qi, Li~Yi, Hao Su, and Leonidas~J. Guibas.
\newblock {PointNet++}: Deep hierarchical feature learning on point sets in a
  metric space.
\newblock In \emph{Advances in Neural Information Processing Systems},
  volume~30, pages 5105--5114, 2017.

\bibitem[Qian et~al.(2022)Qian, Li, Peng, Mai, Hammoud, Elhoseiny, and
  Ghanem]{pointnext}
Guocheng Qian, Yuchen Li, Houwen Peng, Jinjie Mai, Hasan Hammoud, Mohamed
  Elhoseiny, and Bernard Ghanem.
\newblock Pointnext: Revisiting pointnet++ with improved training and scaling
  strategies.
\newblock In S.~Koyejo, S.~Mohamed, A.~Agarwal, D.~Belgrave, K.~Cho, and A.~Oh,
  editors, \emph{Advances in Neural Information Processing Systems}, volume~35,
  pages 23192--23204. Curran Associates, Inc., 2022.
\newblock URL
  \url{https://proceedings.neurips.cc/paper_files/paper/2022/file/9318763d049edf9a1f2779b2a59911d3-Paper-Conference.pdf}.

\bibitem[Qin et~al.(2024)Qin, Li, Zhou, Wang, and Pfister]{property-lang2}
Minghan Qin, Wanhua Li, Jiawei Zhou, Haoqian Wang, and Hanspeter Pfister.
\newblock Langsplat: 3d language gaussian splatting.
\newblock In \emph{2024 IEEE/CVF Conference on Computer Vision and Pattern
  Recognition (CVPR)}, pages 20051--20060, 2024.

\bibitem[Radford et~al.(2021)Radford, Kim, Hallacy, Ramesh, Goh, Agarwal,
  Sastry, Askell, Mishkin, Clark, Krueger, and Sutskever]{clip}
Alec Radford, Jong~Wook Kim, Chris Hallacy, Aditya Ramesh, Gabriel Goh,
  Sandhini Agarwal, Girish Sastry, Amanda Askell, Pamela Mishkin, Jack Clark,
  Gretchen Krueger, and Ilya Sutskever.
\newblock Learning transferable visual models from natural language
  supervision.
\newblock In Marina Meila and Tong Zhang, editors, \emph{Proceedings of the
  38th International Conference on Machine Learning}, volume 139 of
  \emph{Proceedings of Machine Learning Research}, pages 8748--8763. PMLR,
  2021.
\newblock URL \url{https://proceedings.mlr.press/v139/radford21a.html}.

\bibitem[Ravi et~al.(2025)Ravi, Gabeur, Hu, Hu, Ryali, Ma, Khedr, Rädle,
  Rolland, Gustafson, Mintun, Pan, Alwala, Carion, Wu, Girshick, Dollár, and
  Feichtenhofer]{sam2}
Nikhila Ravi, Valentin Gabeur, Yuan-Ting Hu, Ronghang Hu, Chaitanya Ryali,
  Tengyu Ma, Haitham Khedr, Roman Rädle, Chloe Rolland, Laura Gustafson, Eric
  Mintun, Junting Pan, Kalyan~Vasudev Alwala, Nicolas Carion, Chao-Yuan Wu,
  Ross Girshick, Piotr Dollár, and Christoph Feichtenhofer.
\newblock {SAM 2}: Segment anything in images and videos.
\newblock In \emph{The Thirteenth International Conference on Learning
  Representations}, 2025.

\bibitem[Shen et~al.(2024)Shen, Yang, and Wang]{flashsplat}
Qiuhong Shen, Xingyi Yang, and Xinchao Wang.
\newblock {FlashSplat}: {2D} to {3D} gaussian splatting segmentation solved
  optimally.
\newblock In \emph{European Conference on Computer Vision}, pages 456--472.
  Springer, 2024.
\newblock \doi{10.1007/978-3-031-72670-5_26}.

\bibitem[Shi et~al.(2024)Shi, Wang, Duan, and Guan]{property-lang}
Jin-Chuan Shi, Miao Wang, Hao-Bin Duan, and Shao-Hua Guan.
\newblock Language embedded 3d gaussians for open-vocabulary scene
  understanding.
\newblock In \emph{2024 IEEE/CVF Conference on Computer Vision and Pattern
  Recognition (CVPR)}, pages 5333--5343, 2024.

\bibitem[Song et~al.(2026)Song, Li, Wang, and Yan]{pointgs}
Yixiao Song, Qingyong Li, Wen Wang, and Zhicheng Yan.
\newblock {PointGS}: Semantic-consistent unsupervised {3D} point cloud
  segmentation with {3D} gaussian splatting.
\newblock In \emph{Proceedings of the IEEE/CVF Conference on Computer Vision
  and Pattern Recognition}, pages 33343--33352, 2026.

\bibitem[Sun et~al.(2025)Sun, Xu, Wu, Zhang, Chen, Ma, Zelek, and
  Li]{pointgauss}
Wentao Sun, Hanqing Xu, Quanyun Wu, Dedong Zhang, Yiping Chen, Lingfei Ma,
  John~S. Zelek, and Jonathan Li.
\newblock {PointGauss}: Point cloud-guided multi-object segmentation for
  gaussian splatting, 2025.
\newblock URL \url{https://arxiv.org/abs/2508.00259}.

\bibitem[Tang et~al.(2022)Tang, Zhan, Chen, Yu, and Tao]{contrastiveboundary}
Liyao Tang, Yibing Zhan, Zhe Chen, Baosheng Yu, and Dacheng Tao.
\newblock Contrastive boundary learning for point cloud segmentation.
\newblock In \emph{Proceedings of the IEEE/CVF conference on computer vision
  and pattern recognition}, pages 8489--8499, 2022.

\bibitem[Wu et~al.(2022)Wu, Lao, Jiang, Liu, and Zhao]{point4}
Xiaoyang Wu, Yixing Lao, Li~Jiang, Xihui Liu, and Hengshuang Zhao.
\newblock Point transformer v2: Grouped vector attention and partition-based
  pooling.
\newblock In \emph{Advances in Neural Information Processing Systems},
  volume~35, pages 33330--33342, 2022.

\bibitem[Wu et~al.(2024)Wu, Jiang, Wang, Liu, Liu, Qiao, Ouyang, He, and
  Zhao]{point2}
Xiaoyang Wu, Li~Jiang, Peng-Shuai Wang, Zhijian Liu, Xihui Liu, Yu~Qiao, Wanli
  Ouyang, Tong He, and Hengshuang Zhao.
\newblock Point transformer v3: Simpler faster stronger.
\newblock In \emph{Proceedings of the IEEE/CVF Conference on Computer Vision
  and Pattern Recognition}, pages 4840--4851, 2024.

\bibitem[Xie et~al.(2020)Xie, Gu, Guo, Qi, Guibas, and Litany]{pointcontrast}
Saining Xie, Jiatao Gu, Demi Guo, Charles~R Qi, Leonidas Guibas, and Or~Litany.
\newblock Pointcontrast: Unsupervised pre-training for 3d point cloud
  understanding.
\newblock In \emph{European conference on computer vision}, pages 574--591.
  Springer, 2020.

\bibitem[Xu et~al.(2024)Xu, Yuan, Li, Zhang, Jie, Ma, Tang, Sebe, and
  Wang]{3DSS-VLG}
Xiaoxu Xu, Yitian Yuan, Jinlong Li, Qiudan Zhang, Zequn Jie, Lin Ma, Hao Tang,
  Nicu Sebe, and Xu~Wang.
\newblock 3d weakly supervised semantic segmentation with 2d vision-language
  guidance.
\newblock In \emph{European Conference on Computer Vision}, pages 87--104.
  Springer, 2024.

\bibitem[Ye et~al.(2024)Ye, Danelljan, Yu, and Ke]{gaussian_grouping}
Mingqiao Ye, Martin Danelljan, Fisher Yu, and Lei Ke.
\newblock Gaussian {G}rouping: Segment and edit anything in {3D} scenes.
\newblock In \emph{Proceedings of the European Conference on Computer Vision
  (ECCV)}, pages 162--179, 2024.

\bibitem[Zhou et~al.(2024)Zhou, Chang, Jiang, Fan, Zhu, Xu, Chari, You, Wang,
  and Kadambi]{feature3dgs}
Shijie Zhou, Haoran Chang, Sicheng Jiang, Zhiwen Fan, Zehao Zhu, Dejia Xu,
  Pradyumna Chari, Suya You, Zhangyang Wang, and Achuta Kadambi.
\newblock Feature {3DGS}: Supercharging {3D} gaussian splatting to enable
  distilled feature fields.
\newblock In \emph{Proceedings of the IEEE/CVF Conference on Computer Vision
  and Pattern Recognition (CVPR)}, pages 21676--21685, 2024.

\end{thebibliography}

\clearpage
\appendix
\setcounter{figure}{0}
\setcounter{table}{0}
\renewcommand{\thefigure}{S\arabic{figure}}
\renewcommand{\thetable}{S\arabic{table}}
\makeatletter
\setlength{\@dblfptop}{0pt}
\setlength{\@dblfpsep}{12pt}
\setlength{\@dblfpbot}{0pt plus 1fil}
\makeatother
\section*{Supplementary Material}
\section{Supplementary Overview}

This supplement reports the evaluated scenes, implementation and timing boundaries, per-window KITTI-360 results, the complete ScanNet class breakdown, the full view-coverage ablation, and examples of mapping the completed 3D labels back to image views. DesktopObjects-360 and NeRDS-360 use optimized Gaussian scenes (Mode II); ScanNet and KITTI-360 use one point-completed Gaussian per evaluation point (Mode I).

\section{Datasets and Evaluated Scenes}

\paragraph{DesktopObjects-360.}
DesktopObjects-360 was introduced with PointGauss~\cite{pointgauss}. It provides annotated point clouds, multi-view images, and pretrained Gaussian Splatting models for six tabletop scenes with 7--10 objects per scene, totaling 3,364 images and 56 annotated 3D instances. We use the released optimized Gaussian models and evaluate prompted object segmentation.

\paragraph{NeRDS-360.}
NeRDS-360~\cite{nerds360} contains outdoor scenes reconstructed from COLMAP inputs with approximately 360-degree camera coverage. For each of the ten urban scenes evaluated here, we independently optimize a 2DGS~\cite{2dgs} representation for 30,000 iterations using the original calibrated images. We manually annotate the vehicle instances on the evaluation geometry; these labels are withheld from CDSeg and used only for metric computation.

\paragraph{KITTI-360.}
KITTI-360~\cite{kitti360} contains driving images, LiDAR scans, calibrated poses, and point-level semantic and instance labels. We extract ten static windows from sequence 0 and retain stationary vehicles as targets. This avoids assigning a single static Gaussian carrier to objects that move during the window.

\paragraph{ScanNet.}
ScanNet~\cite{scannet} provides indoor RGB-D sequences, reconstructed geometry, camera poses, and semantic annotations. We use it to evaluate semantic-mask association independently of image recognition. The precise geometry and mask protocol are described in the \emph{ScanNet Protocol and Class Results} section.

\paragraph{Choice of Gaussian carrier.}
The carrier is selected according to the available 3D representation and the evaluation space. DesktopObjects-360 is evaluated on the optimized Gaussian scenes released with PointGauss~\cite{pointgauss}, whereas NeRDS-360 uses our independently optimized 2DGS scenes; both use Mode II. KITTI-360 and ScanNet labels are tied to points, so Gaussian completion (Mode I) preserves a one-to-one index between each input point and its completed Gaussian. This assignment is fixed for each dataset and is used by all reported comparisons. Mode I therefore returns labels in input-point order, whereas Mode II retains labels on the native Gaussian primitives.

\begin{table*}[t]
\centering
{
\small
\begin{tabular}{@{}lrrlrr@{}}
\toprule
Scene & Points & Images & Scene & Points & Images \\
\midrule
\multicolumn{3}{c}{\textbf{DesktopObjects-360}} & \multicolumn{3}{c}{\textbf{NeRDS-360}} \\
Desk1 & 154,947 & 324 & 6thAndMission-medium0 & 395,517 & 198 \\
Desk2 & 121,152 & 373 & 6thAndMission-medium10 & 435,532 & 199 \\
Desk3 & 174,916 & 957 & 6thAndMission-medium12 & 425,399 & 198 \\
Desk4 & 138,378 & 740 & 6thAndMission-medium6 & 581,460 & 192 \\
Desk5 & 192,247 & 764 & 6thAndMission-medium7 & 340,734 & 199 \\
Desk6 & 119,210 & 206 & GrantAndCalifornia1 & 356,313 & 199 \\
 & & & GrantAndCalifornia2 & 396,088 & 195 \\
 & & & GrantAndCalifornia3 & 337,053 & 196 \\
 & & & VanNessAveAndTurkSt3 & 336,844 & 192 \\
 & & & VanNessAveAndTurkSt5 & 482,906 & 199 \\
\bottomrule
\end{tabular}
}
\caption{DesktopObjects-360 and NeRDS-360 scenes used in the experiments. Counts refer to the evaluated 3D representation and calibrated input images.}
\label{tab:sup_scene_sizes}
\end{table*}

\begin{table*}[t]
\centering
{
\small
\begin{tabular}{@{}lrrlrr@{}}
\toprule
KITTI-360 window & Points & Images & ScanNet scene & Points & Images \\
\midrule
372--610 & 3,063,621 & 239 & scene0000\_00 & 1,990,518 & 300 \\
599--846 & 3,503,785 & 248 & scene0002\_00 & 1,512,778 & 300 \\
834--1286 & 3,609,825 & 381 & scene0024\_00 & 6,642,655 & 300 \\
1270--1549 & 3,365,820 & 280 & scene0030\_00 & 7,185,251 & 300 \\
1537--1755 & 2,476,010 & 219 & & & \\
1740--1991 & 3,007,107 & 252 & & & \\
2695--2925 & 2,917,051 & 231 & & & \\
3221--3475 & 2,980,566 & 255 & & & \\
3463--3724 & 3,117,777 & 262 & & & \\
3711--3928 & 2,966,142 & 218 & & & \\
\bottomrule
\end{tabular}
}
\caption{KITTI-360 sequence-0 windows and representative ScanNet scenes. ScanNet counts describe the dense visualization geometry; semantic metrics use all official benchmark vertices.}
\label{tab:sup_point_scene_sizes}
\end{table*}

\section{Implementation and Evaluation Protocol}

All experiments run on a system with an NVIDIA RTX 4090 GPU, an Intel i5-11400F CPU, and 32 GB of RAM, using PyTorch 2.5.1 and CUDA 12.4. For Mode I, each point retains its original index after Gaussian completion; no densification, pruning, or radiance-field optimization is performed. The local $k$-nearest-neighbor filter uses $k=3$. A primitive receiving no valid image observation is assigned label 0, which is treated as background. If two or more labels receive the same number of votes, the smaller numerical label is selected; the same deterministic rule is used for a tie in the neighborhood filter. For Mode II, the optimized Gaussian scenes are loaded before segmentation: DesktopObjects-360 uses its released models, while each NeRDS-360 model is independently optimized with 2DGS for 30,000 iterations. The manually annotated NeRDS-360 vehicle labels are loaded only for final metric computation.

We use the images at the native resolution and calibrated viewpoints provided by each dataset, without introducing resized inputs or synthesized cameras. Unless the number of views is explicitly varied in the view-coverage experiment, the dataset-provided view set is used. For prompted segmentation, a first-view prompt initializes SAM2~\cite{sam2} with the \texttt{sam2.1\_hiera\_small.pt} checkpoint; cameras are ordered by center proximity and processed as a pseudo-video. For automatic instances, YOLO11~\cite{yolo11_ultralytics} with the \texttt{yolo11l-seg.pt} checkpoint supplies image masks and tracked instance identities. ScanNet instead uses the provided semantic image annotations, so its numbers isolate the 2D--3D association and fusion stages.

For the reported CDSeg runtime, timing starts after the required images have been read and the input masks, camera parameters, and Gaussian carrier are available to the method. Timing stops when the final filtered 3D labels are returned. It therefore covers renderer-based correspondence extraction, voting, and neighborhood filtering, while excluding disk input, external 2D-mask inference, and prior Gaussian reconstruction. The submitted code archive provides the exact timing calls and the remaining execution details.

The supervised point baselines use the default configurations of their official implementations. On DesktopObjects-360 and NeRDS-360, they follow leave-one-scene-out evaluation. Projection, z-buffer projection, and CDSeg always consume the same image masks. Reported metrics are mean intersection-over-union (mIoU), overall point accuracy (mAcc), and mean per-class accuracy (mAcc-cls); KITTI-360 additionally reports precision, recall, and F1 for stationary vehicles.

\section{Extended KITTI-360 Results}
\label{sec:sup_kitti}

Road scenes contain moving vehicles and heavily occluded structures. Because CDSeg assumes a static 3D carrier within each evaluated window, we select static intervals and evaluate stationary vehicles. Building instances are not used: trees and other foreground objects frequently occlude their image boundaries, and the available masks do not provide reliable building identities across the selected views.

Table~\ref{tab:sup_kitti_results} reports every window. The aggregate mIoU is 57.44\%, with 83.77\% precision and 64.48\% recall. The gap is systematic rather than confined to one scene: image-visible vehicle sides are usually precise, while roofs, front surfaces, and LiDAR points outside the camera field of view remain unlabeled. Window 372--610 obtains the highest mIoU (65.91\%); window 834--1286 is the most difficult (38.07\%) and also has the lowest recall.

\begin{table*}[t]
\centering
{
\small
\begin{tabular}{@{}lrrrrrr@{}}
\toprule
Window & mIoU & mAcc & mAcc-cls & Precision & Recall & F1 \\
\midrule
372--610 & 65.91 & 99.25 & 67.02 & 97.67 & 68.19 & 80.31 \\
599--846 & 60.92 & 98.97 & 67.75 & 86.34 & 67.86 & 75.99 \\
834--1286 & 38.07 & 96.47 & 47.08 & 67.20 & 49.38 & 56.93 \\
1270--1549 & 55.58 & 97.97 & 56.97 & 96.72 & 60.89 & 74.73 \\
1537--1755 & 64.57 & 97.62 & 74.34 & 81.47 & 74.52 & 77.84 \\
1740--1991 & 60.28 & 98.50 & 65.37 & 88.94 & 67.70 & 76.88 \\
2695--2925 & 52.32 & 98.32 & 56.98 & 83.95 & 57.02 & 67.92 \\
3221--3475 & 54.68 & 97.66 & 66.84 & 75.91 & 71.39 & 73.58 \\
3463--3724 & 57.17 & 98.55 & 58.99 & 96.00 & 64.52 & 77.17 \\
3711--3928 & 64.91 & 98.85 & 72.91 & 80.40 & 72.52 & 76.26 \\
\midrule
All & 57.44 & 98.22 & 63.42 & 83.77 & 64.48 & 72.87 \\
\bottomrule
\end{tabular}
}
\caption{Stationary-vehicle instance segmentation on the ten KITTI-360 windows. All values are percentages.}
\label{tab:sup_kitti_results}
\end{table*}

\section{ScanNet Protocol and Class Results}
\label{sec:sup_scannet}

We evaluate on the full ScanNet-v2 validation split and use the vertices of the official semantic benchmark mesh, \texttt{scene\_vh\_clean\_2.ply}. The corresponding \texttt{scene\_vh\_clean\_2.labels.ply} files contain the same geometry and the evaluation labels. CDSeg reads only vertex coordinates and appearance when constructing the carrier; the semantic label property is withheld until metric computation. The provided semantic image annotations, which have relatively coarse object boundaries, remain the image-domain input, so the experiment measures correspondence and multi-view fusion without training a 3D segmentation network.

Following the benchmark convention, we accumulate one confusion matrix over the entire validation split, ignore label 0 and other non-evaluation labels, compute each of the 20 class IoUs from that global matrix, and then average them. This gives the standard mIoU reported in the main paper. The evaluation contains 35,881,054 valid ground-truth vertices. We additionally report global valid-point accuracy. Scene-averaged quantities are retained only as diagnostics and are not used in comparisons with prior work.

\begin{figure*}[t]
\centering
\includegraphics[width=\linewidth]{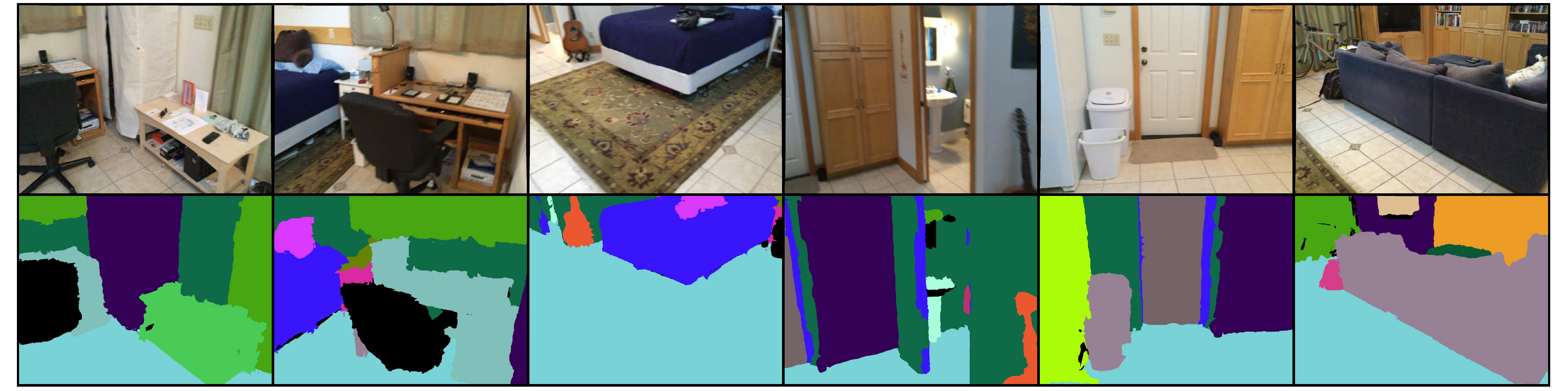}
\caption{Representative ScanNet RGB views (top) and the provided semantic annotations used as image-domain inputs (bottom).}
\label{fig:sup_scannet_masks}
\end{figure*}

Figure~\ref{fig:sup_scannet_masks} shows representative RGB frames and input masks. Boundaries in the image annotations and reconstructed point geometry are not perfectly aligned, which creates contradictory labels near object contours even without semantic recognition errors. Completed scenes and enlarged furniture regions are shown in the main paper.

CDSeg obtains 65.7749\% benchmark mIoU and 73.3706\% global valid-point accuracy. Table~\ref{tab:sup_scannet_classes} reports the per-class IoU averaged across scenes; the macro-average of these 20 diagnostic values is 63.881\%. This differs from benchmark mIoU because the benchmark first accumulates one global confusion matrix over all valid vertices. Bathtub reaches 77.43\% average IoU, while wall and floor reach 74.93\% and 74.04\%, demonstrating strong transfer across both object and structural categories.

\begin{table}[t]
\centering
{
\small
\begin{tabular}{@{}lrlr@{}}
\toprule
\multicolumn{4}{c}{Per-class macro-average: 63.881\%} \\
\midrule
Class & IoU & Class & IoU \\
\midrule
wall & 74.93 & floor & 74.04 \\
cabinet & 56.04 & bed & 58.54 \\
chair & 54.25 & sofa & 67.48 \\
table & 55.18 & door & 64.57 \\
window & 66.64 & bookshelf & 63.06 \\
picture & 71.64 & counter & 59.07 \\
desk & 57.49 & curtain & 68.17 \\
refrigerator & 64.59 & shower curtain & 63.24 \\
toilet & 62.33 & sink & 66.91 \\
bathtub & 77.43 & other furniture & 52.02 \\
\bottomrule
\end{tabular}
}
\caption{Scene-averaged per-class IoU on the ScanNet-v2 validation split. These diagnostic values are not used in benchmark comparisons.}
\label{tab:sup_scannet_classes}
\end{table}

\section{Complete View-Coverage Ablation}

We progressively reduce the input views on DesktopObjects-360 Desk6. The subsets are selected from the available camera sequence rather than generated by perturbing the masks. Table~\ref{tab:sup_view_ablation} gives the complete results summarized in the main paper. Accuracy remains near 93\% mIoU from 206 down to 10 views when the selected cameras still cover the objects. It decreases to 89.53\% with two views and 74.50\% with one view.

The small nonmonotonic variations between larger subsets show that the identity of the retained cameras matters in addition to their number. A smaller but well-distributed subset may observe the target surfaces better than a larger set containing redundant viewpoints. The pronounced decline at one or two views therefore supports coverage, rather than a fixed minimum view count, as the main bottleneck.

\begin{figure*}[!t]
\centering
\includegraphics[width=\linewidth]{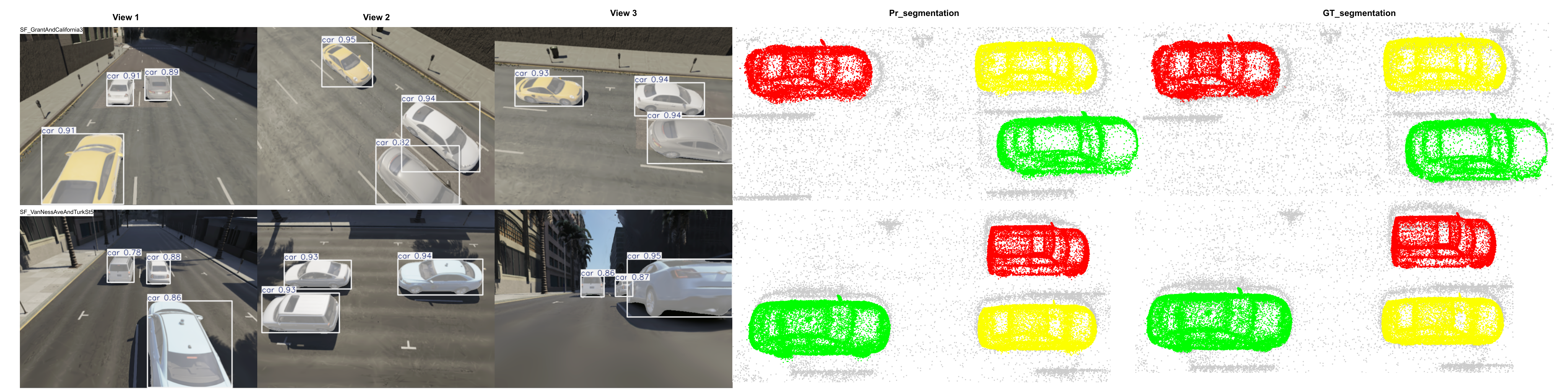}
\caption{Additional NeRDS-360 results. Three input views with automatic vehicle masks are followed by the CDSeg instance prediction and the corresponding ground truth. Rows show two different urban scenes.}
\label{fig:sup_nerds_instances}
\end{figure*}

\begin{table}[t]
\centering
{
\small
\begin{tabular}{@{}rrrr@{}}
\toprule
Views & mIoU & mAcc & mAcc-cls \\
\midrule
206 & 93.36 & 98.88 & 99.17 \\
185 & 93.21 & 98.86 & 99.10 \\
164 & 93.22 & 98.86 & 99.01 \\
144 & 93.07 & 98.84 & 98.64 \\
123 & 92.11 & 98.69 & 97.19 \\
103 & 93.41 & 98.90 & 99.04 \\
82 & 93.20 & 98.86 & 98.94 \\
61 & 93.17 & 98.86 & 98.88 \\
41 & 93.39 & 98.90 & 98.99 \\
20 & 93.17 & 98.87 & 98.77 \\
10 & 93.21 & 98.89 & 98.58 \\
2 & 89.53 & 98.34 & 93.14 \\
1 & 74.50 & 95.80 & 76.08 \\
\bottomrule
\end{tabular}
}
\caption{Full view-count ablation on DesktopObjects-360 Desk6 (\%).}
\label{tab:sup_view_ablation}
\end{table}

\FloatBarrier

\section{Additional NeRDS-360 Qualitative Results}

Figure~\ref{fig:sup_nerds_instances} shows two outdoor scenes not included in the main paper. Three input views are followed by the lifted 3D instance labels and ground truth. Repeated colors denote identities within a scene, not semantic subclasses.

\FloatBarrier

\begin{figure*}[!t]
\centering
\includegraphics[width=\linewidth]{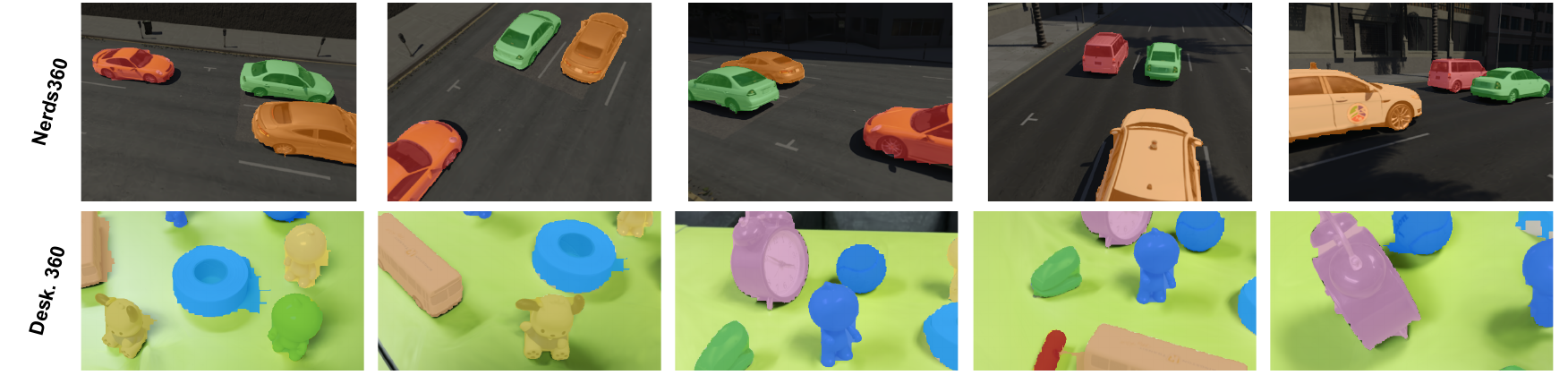}
\caption{Mapping CDSeg's completed 3D instance labels to calibrated image viewpoints. Top: NeRDS-360. Bottom: DesktopObjects-360 (Desk. 360). Colors denote instance identities, and each physical object retains the same color and ID across viewpoints.}
\label{fig:sup_3d2d}
\end{figure*}

\section{Mapping 3D Labels to Images}
\label{sec:sup_3d2d}

Once CDSeg has produced the completed 3D instance labels, they can be returned to calibrated image views by direct projection or by the 3DGS rendering pipeline described in the main paper. For the latter, the renderer monitors accumulated transmittance along each pixel ray. The surface primitive is the first depth-ordered Gaussian after which the remaining transmittance falls below the stopping threshold $\tau_{\mathrm{stop}}=0.1$; its discrete 3D label is then assigned to the pixel. This rule follows the same visibility and occlusion ordering as rendering and does not select the Gaussian with the largest individual compositing weight.

Figure~\ref{fig:sup_3d2d} visualizes this optional image-space output. The upper row contains five viewpoints from NeRDS-360, and the lower row contains five viewpoints from DesktopObjects-360 (Desk. 360). Different colors denote different instance identities. Because every view queries the same completed 3D labeling, each physical object retains its instance ID despite changes in projected shape, scale, and visibility. The results therefore demonstrate 3D-consistent instance identities across image viewpoints; producing continuous 2D multi-view segmentation is not a separate objective of CDSeg.

\section{Detailed Scaling Measurements}

The main paper reports representative points from two scaling tests. Table~\ref{tab:sup_scaling_detail} provides every measurement. Point-count scaling fixes the input at 300 masks, whereas view-count scaling fixes the carrier at one million primitives. Time covers correspondence extraction, voting, and filtering; GPU and CPU values are peak allocated memory.

Across one to seven million primitives, runtime increases from 2.23 to 3.74 seconds and GPU memory from 5.94 to 10.40 GB; CPU memory changes comparatively little. With one million primitives, increasing the number of masks from 10 to 500 raises runtime from 0.21 to 3.30 seconds. Both tests are close to linear over the measured range, consistent with processing each retained view and its visible primitives without materializing a dense primitive--pixel tensor.

\begin{table*}[!t]
\centering
{
\small
\begin{tabular}{@{}rrrr@{\qquad}rrrr@{}}
\toprule
Points (M) & Time (ms) & GPU (MB) & RAM (MB) & Views & Time (ms) & GPU (MB) & RAM (MB) \\
\midrule
1 & 2227.55 & 5944.90 & 5413.34 & 10 & 207.91 & 1159.56 & 1588.14 \\
2 & 2310.34 & 6615.92 & 5498.68 & 50 & 392.77 & 1897.32 & 2053.18 \\
3 & 2553.16 & 7299.68 & 5615.29 & 100 & 890.39 & 2717.10 & 2774.39 \\
4 & 2758.69 & 8053.81 & 5777.40 & 200 & 1496.36 & 4337.86 & 4055.95 \\
5 & 2901.21 & 8804.16 & 5780.20 & 300 & 2227.55 & 5944.90 & 5413.34 \\
6 & 3245.46 & 9537.72 & 5953.02 & 400 & 2791.81 & 7549.86 & 6683.78 \\
7 & 3739.70 & 10400.38 & 5946.12 & 500 & 3295.94 & 9038.75 & 8193.18 \\
\bottomrule
\end{tabular}
}
\caption{Detailed resource measurements. Left: carrier-size scaling with 300 masks. Right: view-count scaling with one million primitives.}
\label{tab:sup_scaling_detail}
\end{table*}





\end{document}